\documentclass[8pt,twocolumn]{article}

\usepackage[a4paper,left=16mm,right=16mm,top=25mm,bottom=25mm]{geometry}

\usepackage{times}
\usepackage{soul}
\usepackage{url}
\usepackage[hidelinks]{hyperref}
\usepackage[utf8]{inputenc}
\usepackage[small]{caption}
\usepackage{graphicx}
\usepackage{amsmath}
\usepackage{amsthm}
\usepackage{booktabs}
\usepackage{algorithm}
\usepackage{algorithmic}
\usepackage[switch]{lineno}
\usepackage{color}
\usepackage{subcaption}

\title{Comparing Task-Agnostic Embedding Models for Tabular Data}
\author{Frederik Hoppe \and Lars Kleinemeier \and Astrid Franz \and Udo Göbel}
\date{CONTACT Software GmbH, Bremen, Germany}

\begin{document}

\maketitle

\begin{abstract}
    Recent foundation models for tabular data achieve strong task-specific performance via in-context learning. Nevertheless, they focus on direct prediction by encapsulating both representation learning and task-specific inference inside a single, resource-intensive network. This work specifically focuses on representation learning, i.e., on transferable, task-agnostic embeddings. We systematically evaluate task-agnostic representations extracted from tabular foundation models (TabPFN, TabICL and TabSTAR) alongside classical feature engineering (TableVectorizer and a sphere model) across a variety of application tasks as outlier detection (ADBench) and supervised learning (TabArena Lite). We find that simple feature engineering methods achieve comparable or superior performance while requiring significantly less computational resources than tabular foundation models. The code is publicly available.\footnote{https://anonymous.4open.science/r/ser34sgf4e5tabds23}
\end{abstract}

\section{Introduction}

Recent foundation models for tabular data, such as TabPFN \cite{hollmann2025tabpfn}, TabICL \cite{tabicl} and TabSTAR \cite{arazi2025tabstarf}, leverage in-context learning to achieve strong task-specific performance on structured datasets. These models are trained on a huge number of (synthetic) datasets to capture underlying patterns in tabular data, enabling them to adapt quickly to new datasets through contextual information. However, unlike large language models (LLMs) that produce versatile, task-agnostic embeddings useful across diverse downstream applications - such as similarity search, retrieval, clustering, and few-shot transfer - these tabular models focus primarily on direct prediction rather than generating transferable representations.

Despite the significant progress in task-specific tabular learning, there remains a substantial gap in exploring task-agnostic embeddings for tabular data. From both research and industrial perspectives, such embeddings offer valuable flexibility by producing representations that can be reused across multiple tasks - including outlier detection, classification and regression - without task-specific customization or retraining. Yet, to the best of the authors' knowledge, there has not been a comprehensive evaluation comparing task-agnostic embedding models for tabular data.

This paper addresses this gap by systematically evaluating task-agnostic embeddings, enabling tasks like similarity search, outlier detection, regression and classification on a single table. They are a step towards applications such as cross-dataset search in data lakes, entity resolution, and comparing records across different sources. Therefore, this study focuses on row embeddings since individual records are the basic unit for most machine learning tasks, unlike column or table embeddings. 

We consider two groups of embedding models: On the one hand embeddings extracted from foundation models like TabICL, TabPFN and TabSTAR as task-agnostic representations of individual data samples (table rows), and on the other hand classical feature engineering methods such as TableVectorizer. For the second group, we introduce a simple sphere model to overcome the limitation of a dataset-specific embedding dimension.

Our primary contribution is to initiate discussion about the potential of general-purpose tabular embeddings, drawing parallels to the success of LLM embeddings. By highlighting both promise and current limitations we aim to establish task-agnostic tabular embeddings as a valuable research direction for the community.

\section{Embedding Models}\label{sec:embedding_models}
Learning effective task-agnostic representations of tabular data is challenging due to its heterogeneous nature, combining numerical and categorical features with complex dependencies that vary across datasets. While this initial study focuses on numerical and categorical tabular data, further research should incorporate other modalities such as text, see Section~\ref{sec:conclusion}. We only use models that can be applied to unseen datasets without further training. Thus, we employ five main approaches to obtain per-row tabular embeddings, spanning classical feature engineering to deep learned embeddings: TabICL, TabPFN and TabSTAR as tabular foundation models pre-trained on specific tasks yet capable of generating task-agnostic row embeddings, and TableVectorizer and a sphere model for statistical encoding. 

\paragraph{TabICL.} TabICL \cite{tabicl} is a tabular foundation model that uses in-context learning for classification without parameter updates. The embeddings are extracted from the first stage, where a set transformer processes columns independently to create distribution-aware features. This is followed by a transformer with rotary positional embeddings that models inter-feature dependencies row-by-row. This process produces semantically rich representations via learnable [CLS] tokens. The embedding dimension is 512.

\paragraph{TabPFN.} TabPFNv2 \cite{hollmann2025tabpfn}, an additional foundation model that has been pre-trained on millions of synthetic datasets, employs a two-way attention mechanism that alternates between features and samples. We use three different methods to compute task-agnostic embeddings, each with dimension 192:
\begin{itemize}
    \item \textbf{TabPFNColumns.} Following the tabpfn-extensions approach\footnote{https://github.com/PriorLabs/tabpfn-extensions}, we generate embeddings through feature masking. In this process, each feature is iteratively predicted from the remaining features using the TabPFNClassifier or TabPFNRegressor depending on whether the feature is numerical or categorical. We then extract the representations of each feature that are aggregated into final row embeddings by taking the average per component.
    \item \textbf{TabPFNConstant.} As a more resource-efficient way we artificially add a constant classification target, employ the TabPFNClassifier and extract the final embeddings.
    \item \textbf{TabPFNRandom.} This approach is similar to the prior approach with an artificial target drawn from a standard normal distribution and using the TabPFNRegressor.
\end{itemize}

\paragraph{TabSTAR.} Although TabSTAR \cite{arazi2025tabstarf} is explicitly designed for target-aware representation, we can extract task-agnostic embeddings after the data is processed by the interaction encoder. With a single artificial regression target of 0 the statistical information of one row is represented in a 384-dimensional embedding. TabSTAR includes the semantics of the column names and is even capable of processing longer text with potential beyond categorical and numerical data.

\paragraph{TableVectorizer.} TableVectorizer from the skrub package\footnote{https://github.com/skrub-data/skrub} serves as a preprocessing baseline that converts tabular data into numerical representations through adaptive column-wise encoding, automatically selecting appropriate encoders (one-hot encoding, datetime decomposition, or dimensionality-reduced string encoding) based on each column's characteristics before concatenating features into row embeddings. It is a non-learned, unsupervised method whose embedding dimension depends on the dataset characteristics (ranging from 4 to 166 in our benchmarks).

\paragraph{Sphere Model.} To overcome the limitation of a dataset-specific embedding dimension, we propose to embed all entities occuring in the table on the unit sphere in $d$ dimensions, where $d$ can be arbitrarily chosen. Assume that the table has $n$ rows and $m$ columns. In addition, without restriction we assume that the first $l$ columns are numerical and the last $m-l$ columns are categorical. Ordinal categorical columns possess an intrinsic order and, hence, are treated as numerical. When we refer to categorical columns, we mean nominal categorical columns.

Let us first consider the numerical columns. As numerical columns possess an intrinsic order, we want to transfer this information to the embeddings. The idea is to map every numerical column $1 \leq i \leq l$ to a half of a great circle on the unit sphere. For each numerical column, we independently sample a vector uniformly on the unit sphere, denoted by $u_1, \ldots, u_l$. This vector serves as a starting point of the embedding great circle. Additionally we need a vector perpendicular to $u_i$ describing the direction of the great circle on the unit sphere. We choose this vector $\tilde{u}_i$ as $\tilde{u}_{i,2k}=u_{i,2k+1}$ and $\tilde{u}_{i,2k+1}=-u_{i,2k}$ for $1\leq k \leq \lfloor\frac{d}{2}\rfloor$. If $d$ is an odd number, we set the last component of $\tilde{u}_{i}$ to zero.
We transform every column entry $e_{ij}$, $1\leq j \leq n$, to an angle in the range $[0.1\cdot\pi,0.9\cdot\pi]$ by
\[ \alpha_{ij} = \left( 0.1+0.8\cdot \frac{e_{ij} - \min_{1\leq j\leq n} e_{ij} }{\max_{1\leq j\leq n} e_{ij} -\min_{1\leq j\leq n} e_{ij}}\right)\cdot\pi.
\]
Finally, the embedding reads as 
\[ v^{\text{num}}_{ij} = \sin(\alpha_{ij})\cdot u_i+\cos(\alpha_{ij})\cdot \tilde{u}_i.
\]
Hence, within each numerical column, the order of the corresponding embeddings aligns with the original order of the table entries, as $\alpha_{ij}$ preserves and transfers it.  

As a (nominal) categorical column has no intrinsic order, all categories can be embedded independently. We do this by simply choosing a random uniformly distributed $d$-dimensional unit vector for each category. 
Note, that in high-dimensional spaces two randomly chosen vectors are nearly orthogonal; hence the probability of choosing close embeddings for different categories is nearly zero.

Finally, we construct row embeddings $h_1,\hdots, h_n$ by calculating the sample mean of the entity embeddings.

Besides its simplicity, the embedding method offers a straightforward extension to unseen rows. For new categories, we simply draw an additional random category vector. Numerical values exceeding the original range still align with the original order. We initially do not use the full half circle $[0,\pi]$ to cope with additional data outside the range. If the distance of new values to the original values is large, they could be scaled by a sigmoid function to the angle ranges $(0,0.1\cdot\pi]$ and $[0.9\cdot\pi,\pi)$.

Furthermore, missing data is handled by mapping an empty table cell onto the zero vector, as it is equally far from any other entity embedding vector.

\section{Experiments}
We evaluate the quality of task-agnostic embeddings on two benchmark suites, use several task-specific predictive algorithms and evaluate the performance employing suited evaluation measures.  

\subsection{Datasets}
As benchmark datasets, we use ADBench \cite{adbench} for unsupervised outlier detection tasks and TabArena \cite{tabarena} for supervised classification and regression tasks. From ADBench, we select only datasets labeled as tabular while excluding those from the image category, resulting in a curated collection of 47 real-world tabular datasets. For TabArena, we use the TabArena Lite preset, which evaluates only the first fold of the first repeat for all tasks. This lightweight configuration reduces computational requirements while maintaining dataset diversity across 51 manually curated OpenML datasets. We impose upper bounds of 15,000 samples and 500 features per dataset to keep the computation time in a reasonable range, resulting in 29 datasets for ADBench and 27 for TabArena.

\subsection{Predictive algorithms}

For ADBench datasets the task-agnostic embeddings are evaluated using specialized outlier detection algorithms, i.e. Local Outlier Factor \cite{localoutlier}, Isolation Forest \cite{isolationforest}, and DeepSVDD \cite{deepsvdd}:

\paragraph{Local Outlier Factor.} We utilize scikit-learn's {\tt LocalOutlierFactor} implementation, which computes local density deviations by comparing each sample's density to its neighbors' densities. Higher scores indicate a greater likelihood of outliers. Since the number of neighbors, {\tt n\_neighbors}, is the only parameter influencing the result, we iterate over this parameter, i.e., {\tt n\_neighbors} $\in\{$5,10,15,20,25,30,35,40,45$\}$, and record the optimal score per embedding model and dataset for evaluation.

\paragraph{Isolation Forest.} We employ scikit-learn's {\tt IsolationForest}, which isolates anomalies through recursive random feature partitioning. Outliers require fewer splits for isolation, yielding higher anomaly scores. We iterate over the number of estimators, i.e., {\tt n\_estimators} $\in\{$50,100,150,200,250$\}$, and record the optimal score per embedding model and dataset for evaluation.

\paragraph{Deep Support Vector Data Description (DeepSVDD).} We use PyOD's DeepSVDD implementation in two configurations:
\vspace*{-0.1cm}
\begin{enumerate}
    \item Default parameters from the package with fixed hidden layer architecture (64, 32), and
	\vspace*{-0.1cm}
    \item Dynamically adjusted hidden layers for higher-dimensional embeddings, where layer sizes are computed as decreasing powers of 2 starting from approximately half the embedding dimension (if at least larger than 64) down to 32.
\end{enumerate}
\vspace*{-0.1cm}
Both variants employ the default parameters from the PyOD package for the remaining parameters, i.e., ReLU activation for the hidden layers, sigmoid activation for the output layer, dropout rate of 0.2 and $\ell_2$-regularization with strength 0.1, with training over 200 epochs using batch size 32. The neural network attempts to map most of the embeddings into a hypersphere. Normal data points are mapped to the interior of the hypersphere, whereas anomalous points are mapped to the exterior.

\vspace*{0.2cm}
The supervised tasks in the TabArena datasets are evaluated with K-Nearest Neighbors and Multi-Layer Perceptrons with Optuna\footnote{https://optuna.org}-based hyperparameter optimization (HPO):

\paragraph{K-Nearest Neighbors (KNN).} We employ scikit-learn's {\tt KNeighborsClassifier} / {\tt KNeighborsRegressor} with distance metrics Euclidean and distance-based weighting scheme. For classification tasks, the model returns class probability predictions; for regression, it outputs continuous values. We iterate over the number of neighbors, i.e., {\tt n\_neighbors} $\in\{$5,10,15,20,25,30,35,40,45$\}$, and record the optimal score per embedding model and dataset for evaluation.

\paragraph{Multi-Layer Perceptron (MLP).} We implement a scikit-learn-based MLP with Optuna hyperparameter optimization using Tree-structured Parzen Estimator (TPE) sampling. The search space encompasses: number of hidden layers (1-5 for classification, 1-3 for regression), hidden dimensions (32-512, log-scaled), dropout rates (0.0-0.5), learning rates (0.0001 to 0.01, log-scaled) and batch sizes (16, 32, 64, 128). Models employ StandardScaler normalization, Adam optimization with 250 training epochs and early stopping with 10-epoch patience. We conduct 50 optimization trials with 5-fold cross-validation. For the optimal parameters, we use 1000 training epochs to obtain the final prediction.

\vspace*{0.2cm}
All evaluators inherit from an abstract base class ensuring consistent training and prediction workflows.
Following the respective benchmark protocols, TabArena Lite uses stratified train/test splits with single-fold evaluation, while ADBench outlier detection employs the full dataset approach.

\subsection{Evaluation measures}
We measure the prediction quality by the area under the receiver operating characteristic curve (AUC) for outlier detection and classification, with the one-vs-rest multiclass strategy. The AUC score ranges from 0 to 1 and should be as large as possible. In case of regression tasks, we use the mean absolute percentage error (MAPE). The MAPE score is non-negative and should be as small as possible.

To compare the performance of the embedding models across several predictive algorithms and datasets, we employ a critical difference diagram \cite{CriticalDifference}. For each dataset and algorithm, the performance is ranked, where  a lower rank indicates a better performance. The critical difference diagram plots the embedding models along a horizontal axis, ordered by their average rank. Black horizontal lines group algorithms with statistically non-significant differences in their ranks, as determined by a post-hoc test at significance level $\alpha$~=~0.05. Consequently, the diagram shows statistically equivalent performance groups and highlights significant differences between others.

In addition to the evaluation measures, we assess the performance of the embedding models by the required computation time (preprocessing and embedding computation, without evaluation). We conducted the experiments on a Google Cloud Platform virtual machine equipped with an NVIDIA A100 40GB GPU.

\begin{figure*}[t]
    \centering
    \includegraphics[width=0.8\linewidth]{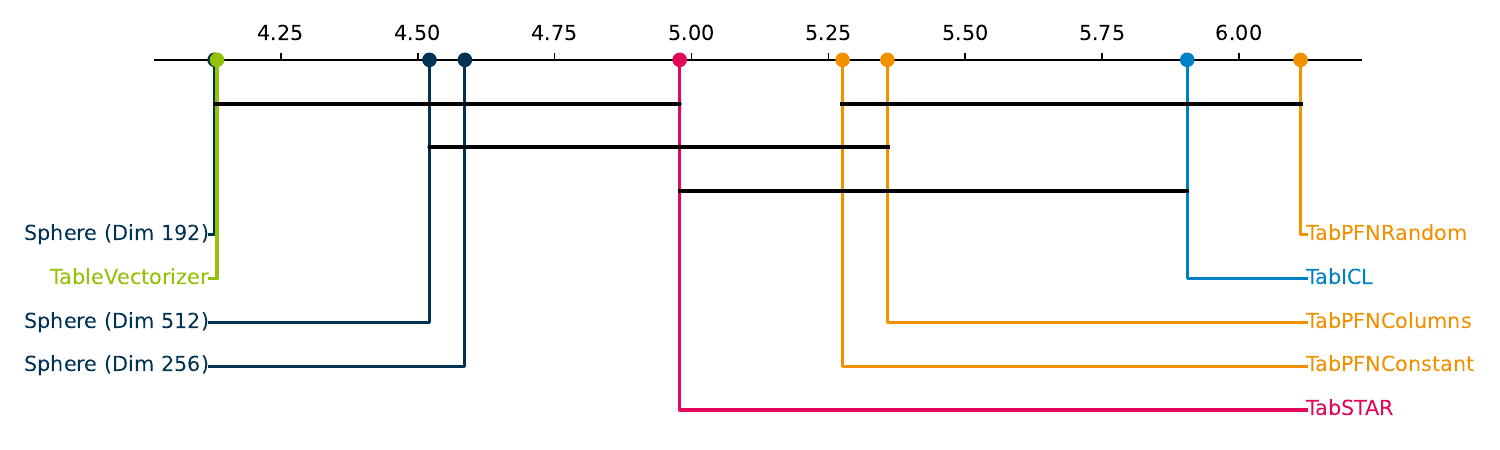}
    \caption{Critical difference diagram for outlier detection.}
    \label{fig:critical_differences_outlier}
\end{figure*}

\begin{figure}
    \centering
    \includegraphics[width=0.8\linewidth]{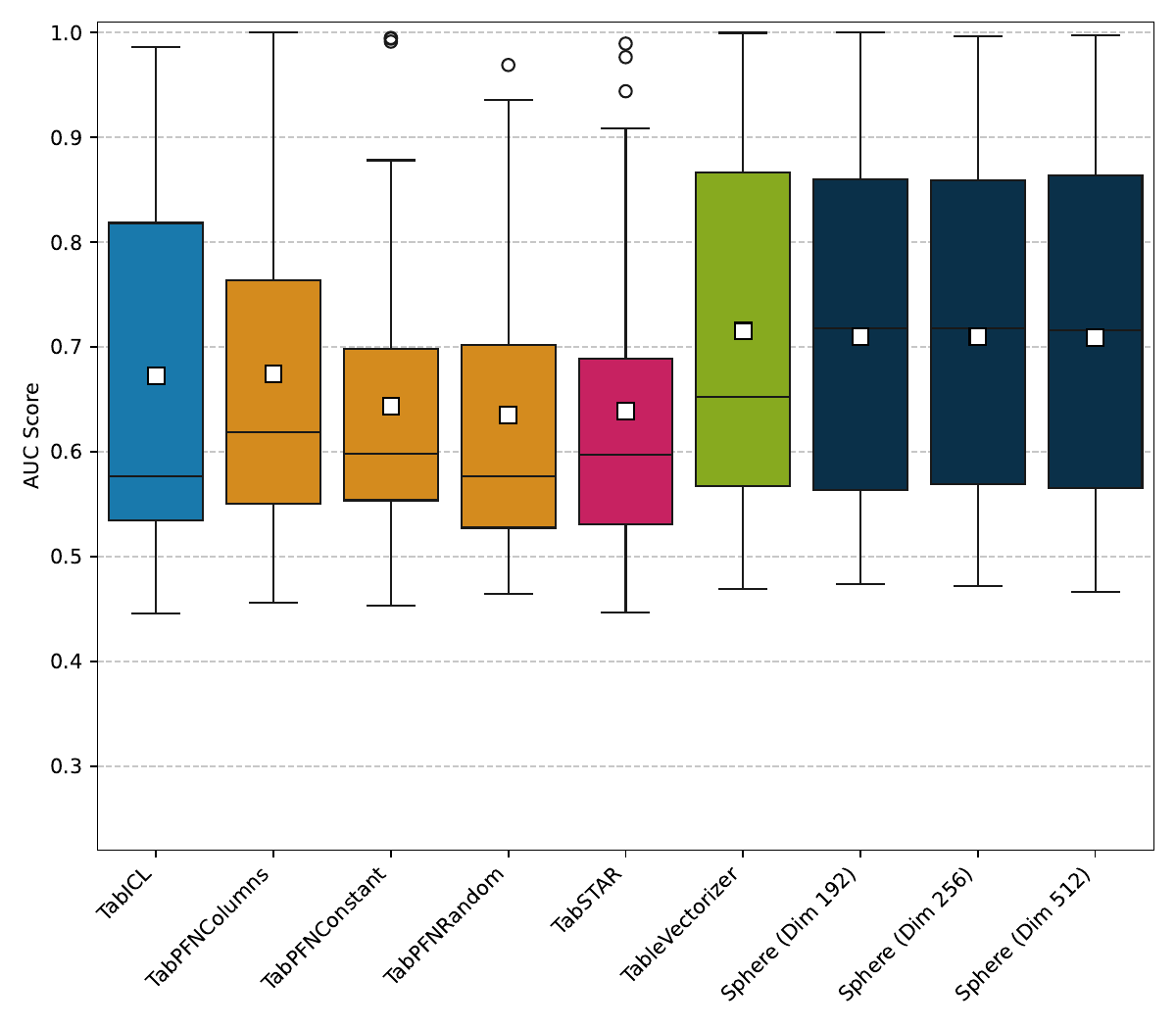}
    \caption{Distribution of the performance for Local Outlier Factor.}
    \label{fig:boxplot_outlier_lof}
\end{figure}

\begin{figure}
    \centering
    \includegraphics[width=0.8\linewidth]{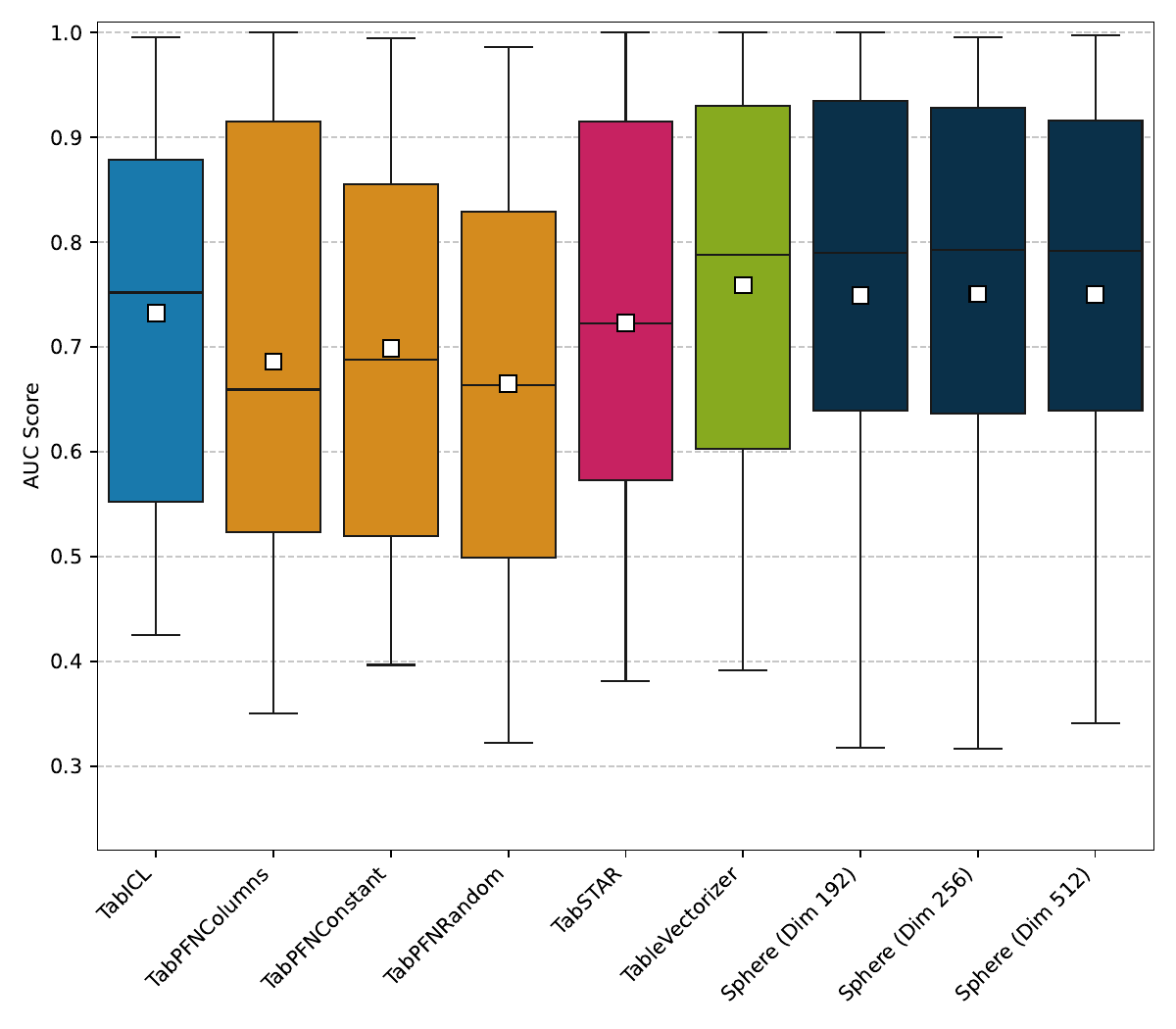}
    \caption{Distribution of the performance for Isolation Forest.}
    \label{fig:boxplot_outlier_isolation_forest}
\end{figure}

\begin{figure}
    \centering
    \includegraphics[width=0.8\linewidth]{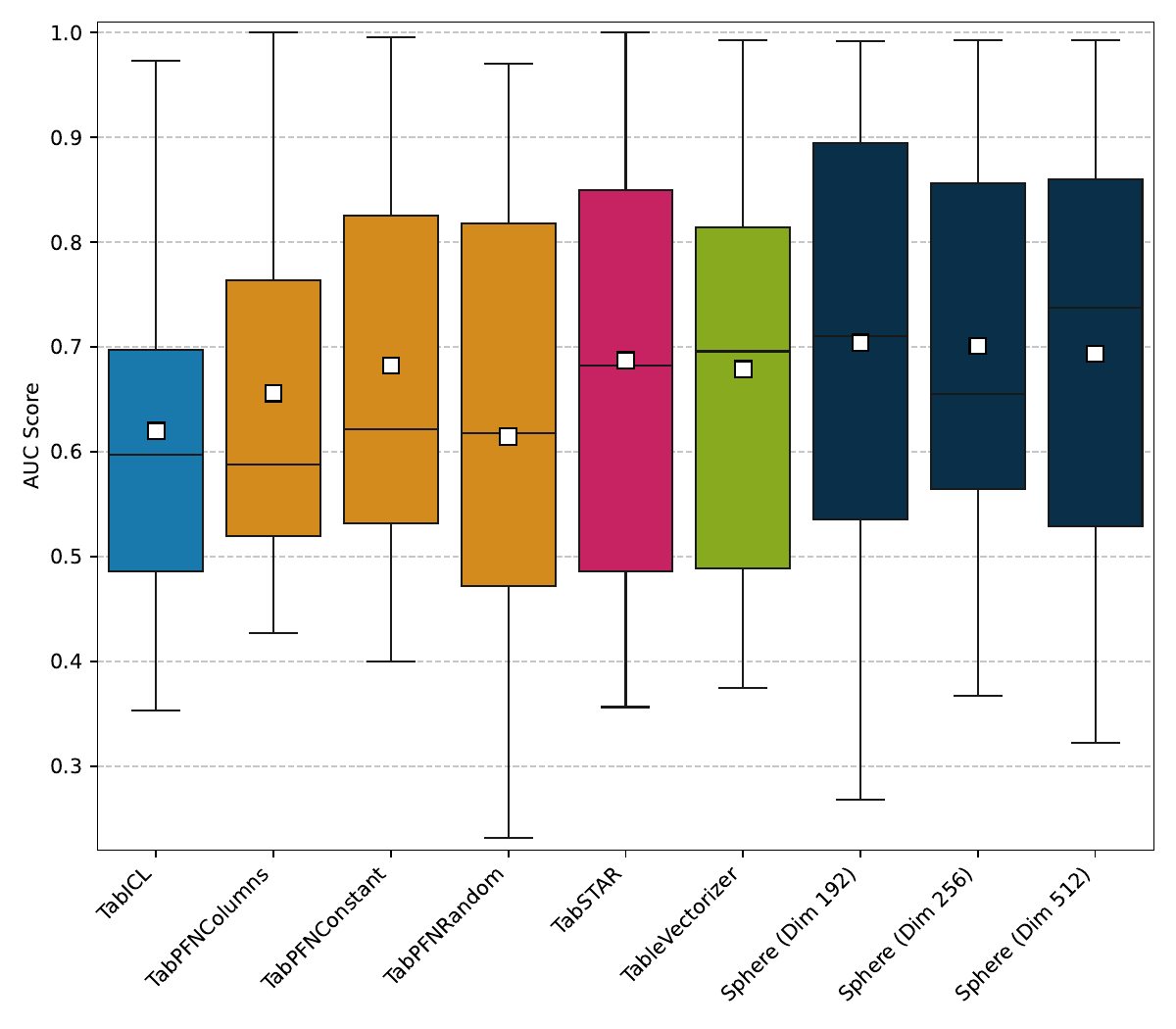}
    \caption{Distribution of the performance for DeepSVDD.}
    \label{fig:boxplot_outlier_deepsvdd}
\end{figure}

\begin{figure}
    \centering
    \includegraphics[width=0.8\linewidth]{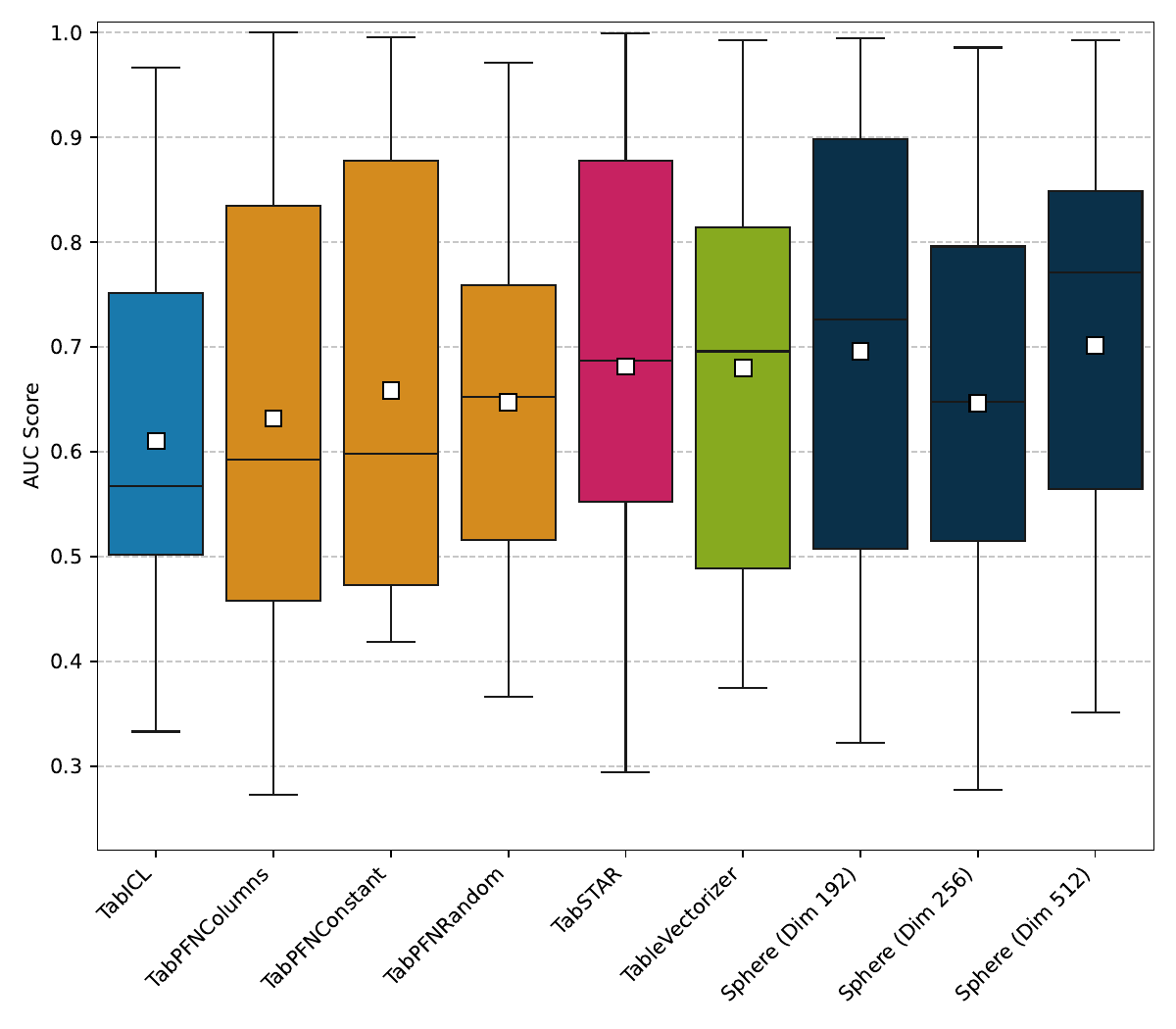}
    \caption{Distribution of the performance for DeepSVDD dynamic.}
    \label{fig:boxplot_outlier_deepsvdd_dynamic}
\end{figure}

\section{Results}

The following discussion focuses on task-agnostic embeddings. Tabular foundation models such as TabICL, TabPFN and TabSTAR are primarily constructed for task-specific prediction leveraging in-context learning. However, they also provide promising task-independent embeddings. Whenever we refer to TabICL, TabPFN and TabSTAR, we mean the task-agnostic embeddings constructed by the specified model as described in Section \ref{sec:embedding_models}. In contrast, we refer to the original end-to-end tabular foundation models as TabICLOriginal and TabPFNOriginal.

Fig.~\ref{fig:critical_differences_outlier} shows the critical difference diagram summarizing the performance over all outlier tasks. From the statistical point of view, TableVectorizer and the sphere model with dimension 192 significantly outperform TabICL and the TabPFN versions. The sphere models, TableVectorizer and TabSTAR on the one hand, and the TabPFN versions alongside TabICL on the other hand are statistically indistinguishable. 

To compare the performance for each of the four outlier detection algorithms separately, we plot the corresponding distribution of AUC scores via boxplots, see Figs.~\ref{fig:boxplot_outlier_lof}-\ref{fig:boxplot_outlier_deepsvdd_dynamic} (median: black line; mean: white square). The observations from Fig.~\ref{fig:critical_differences_outlier} are roughly confirmed on the algorithmic level.

\begin{figure*}[t]
    \centering
    \includegraphics[width=0.8\linewidth]{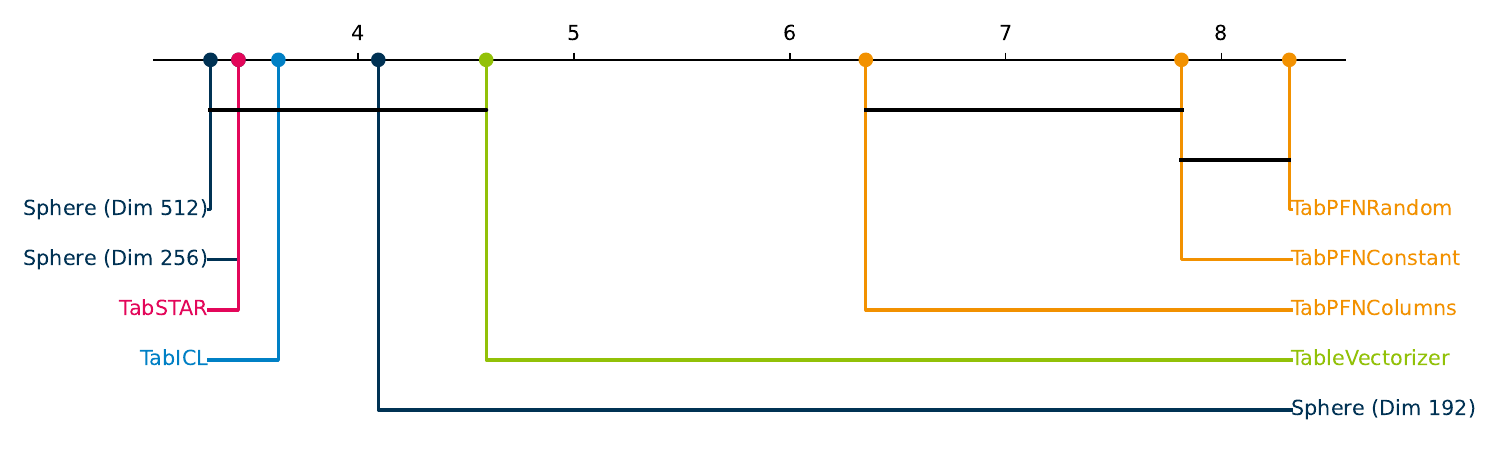}
    \caption{Critical difference diagram for supervised tasks.}
    \label{fig:critical_differences_supervised}
\end{figure*}

The supervised tasks, i.e. binary classification, multiclass classification and regression, are assessed by the critical difference diagram in Fig.~\ref{fig:critical_differences_supervised}. The embedding models cluster in two groups: One group containing the sphere models, TabSTAR, TabICL and TableVectorizer, the other one the TabPFN versions. The former group performs better than the latter, while the models within the first group are statistically indistinguishable. Considering only the rank order, TabSTAR and, more pronounced, TabICL perform better compared to outlier detection.

\begin{figure}[p]
    \centering
    \includegraphics[width=0.8\linewidth]{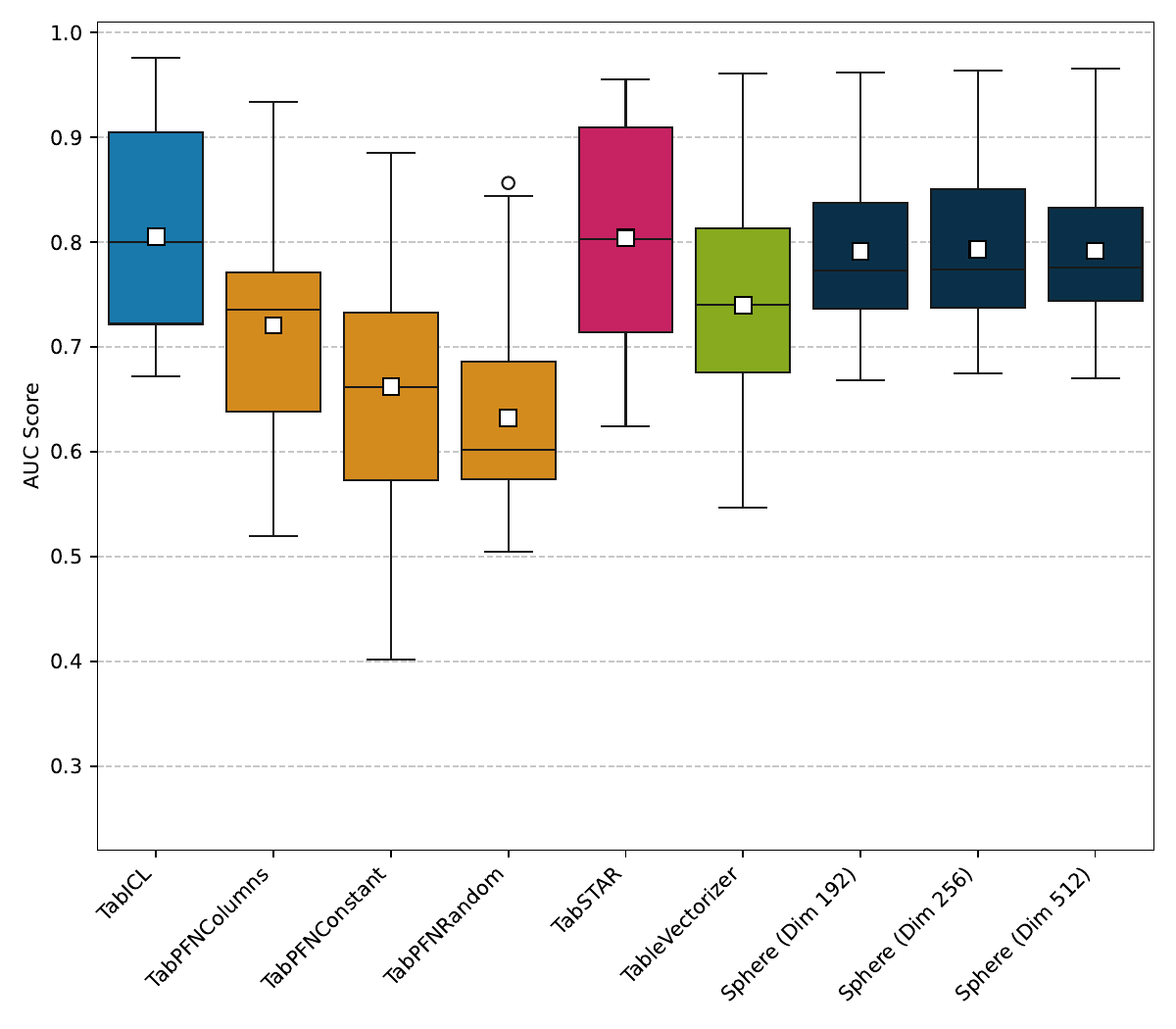}
    \caption{Distribution of the performance for binary classification evaluated via KNN-classifier.}
    \label{fig:boxplot_binary_knn}
\end{figure}

\begin{figure}[p]
    \centering
    \includegraphics[width=0.8\linewidth]{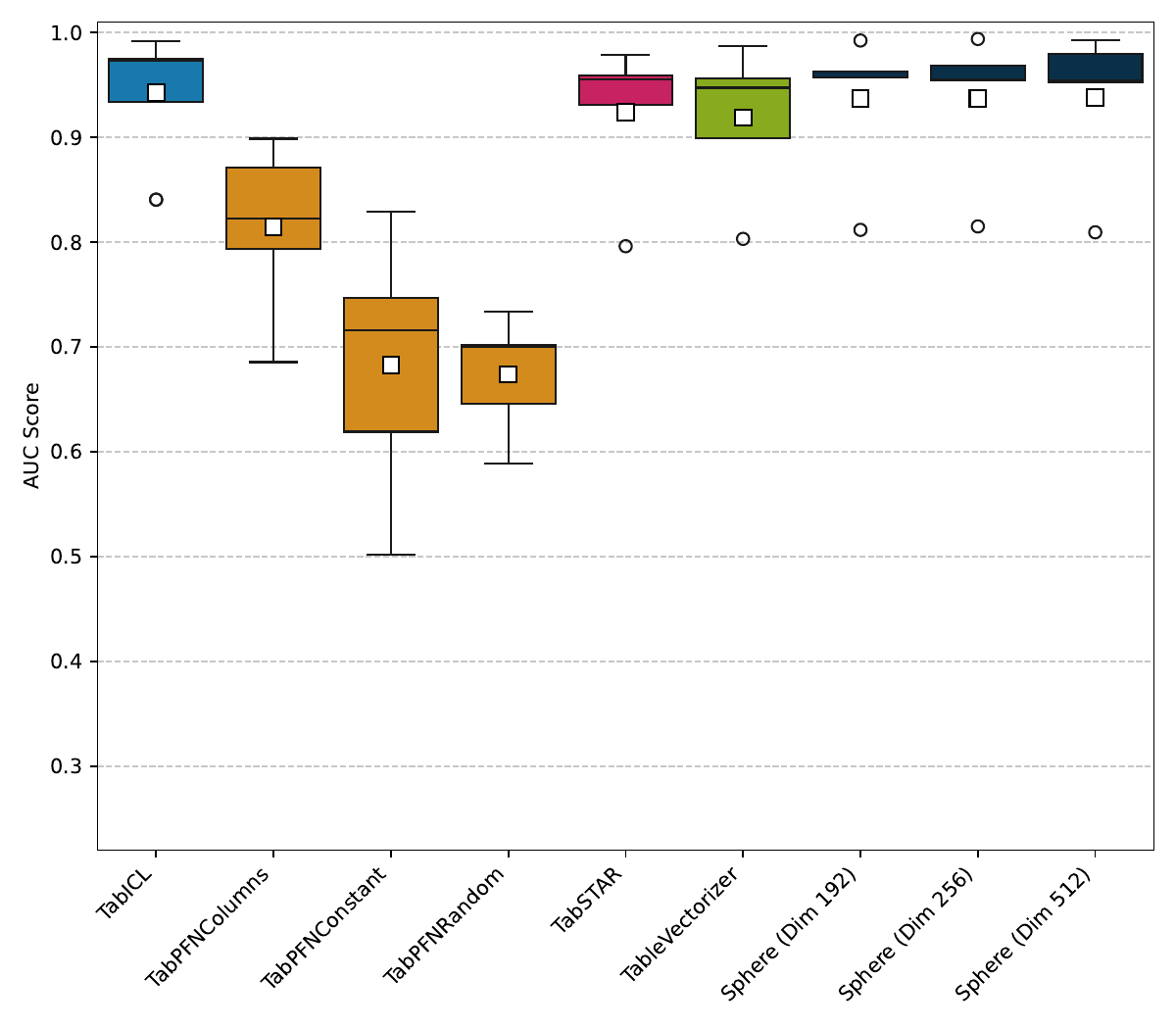}
    \caption{Distribution of the performance for multiclass classification evaluated via KNN-classifier.}
    \label{fig:boxplot_multiclass_knn}
\end{figure}

\begin{figure}[p]
    \centering
    \includegraphics[width=0.8\linewidth]{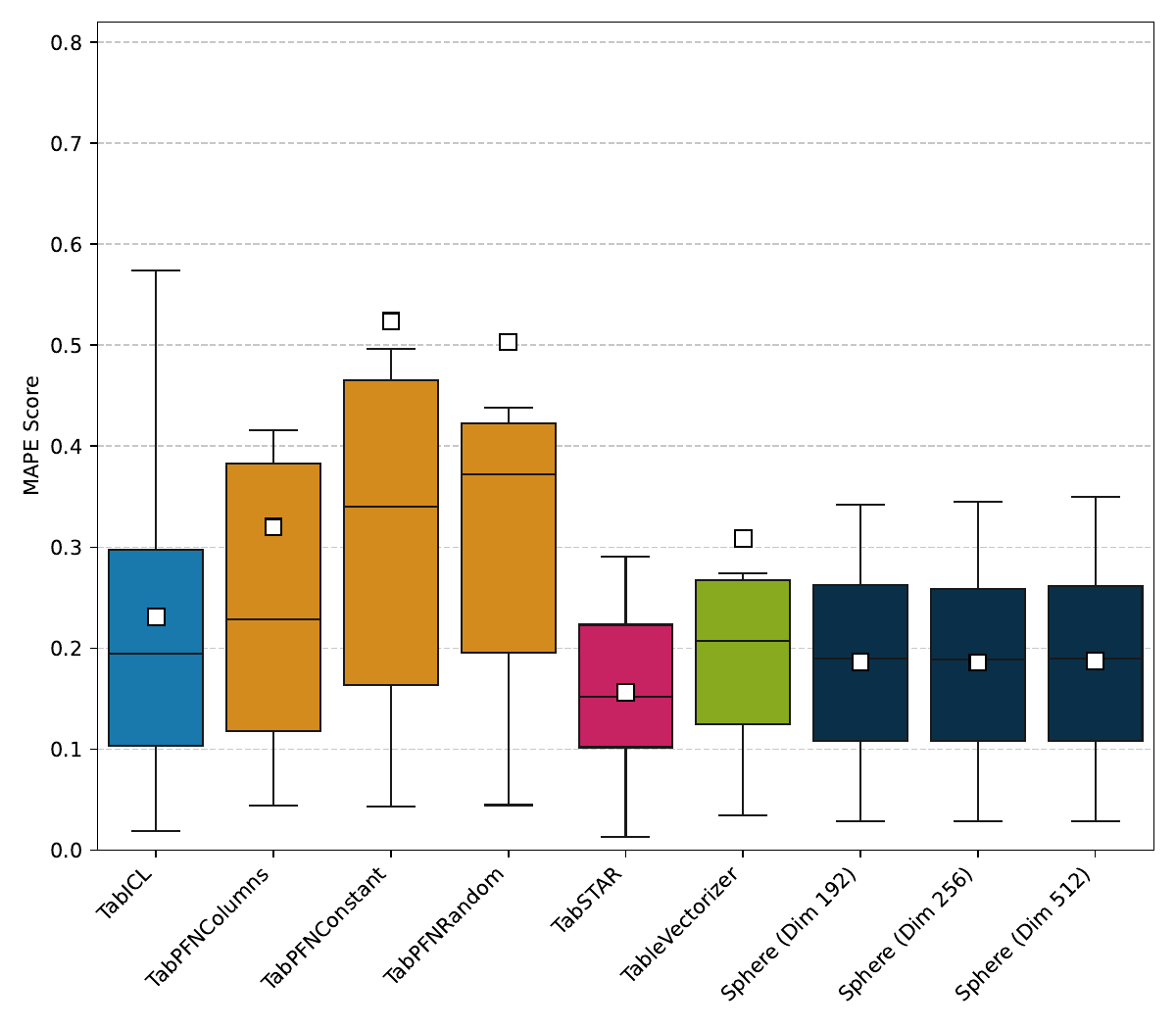}
    \caption{Distribution of the performance for regression via KNN. Note that for TableVectorizer and the TabPFN versions there are outliers that are not captured within the range of the vertical axis.}
    \label{fig:boxplot_regression_knn}
\end{figure}

\begin{figure}[p]
    \centering
    \includegraphics[width=0.8\linewidth]{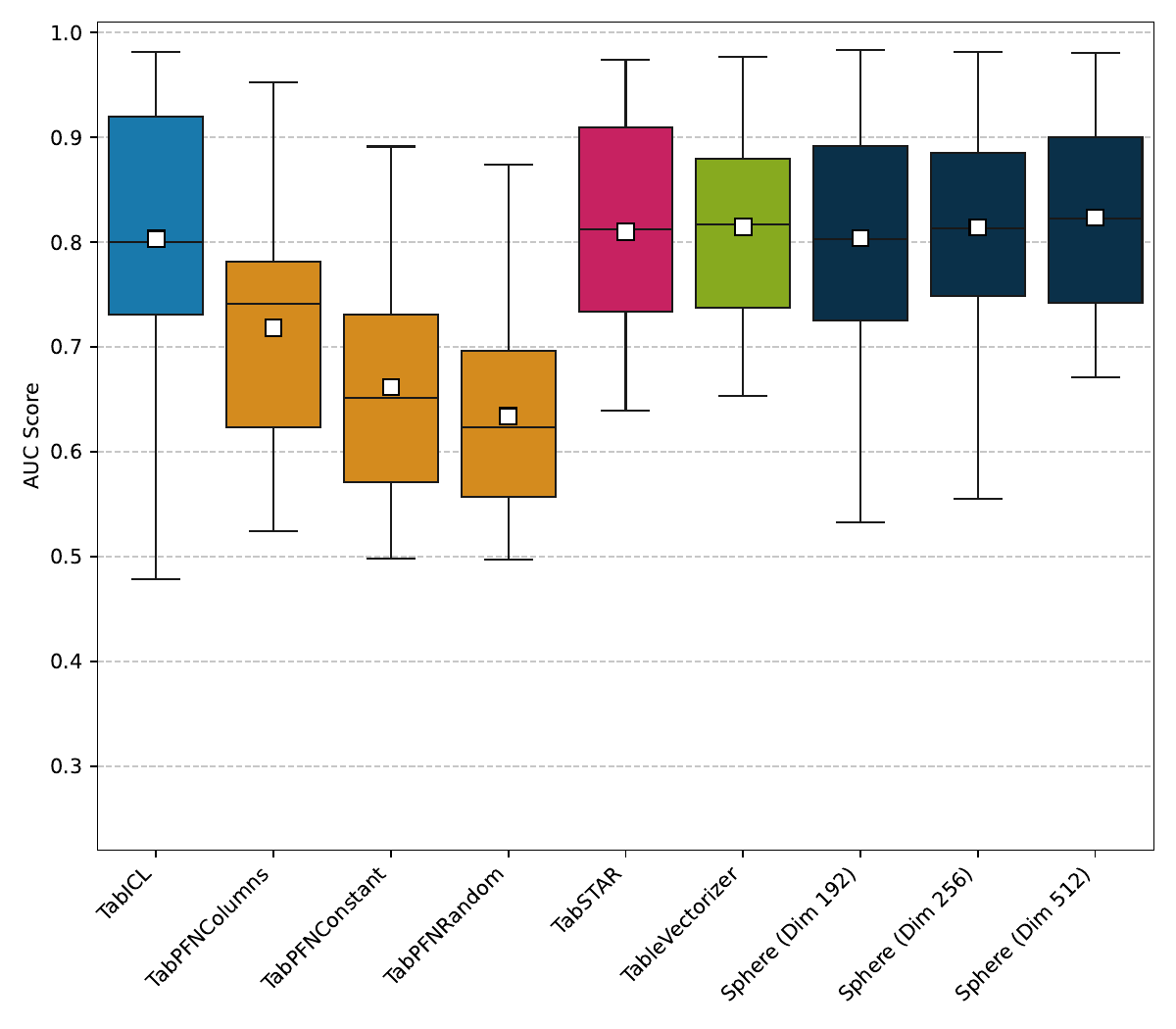}
    \caption{Distribution of the performance for binary classification evaluated via MLP-classifier.}
    \label{fig:boxplot_binary_mlp}
\end{figure}

\begin{figure}[p]
    \centering
    \includegraphics[width=0.8\linewidth]{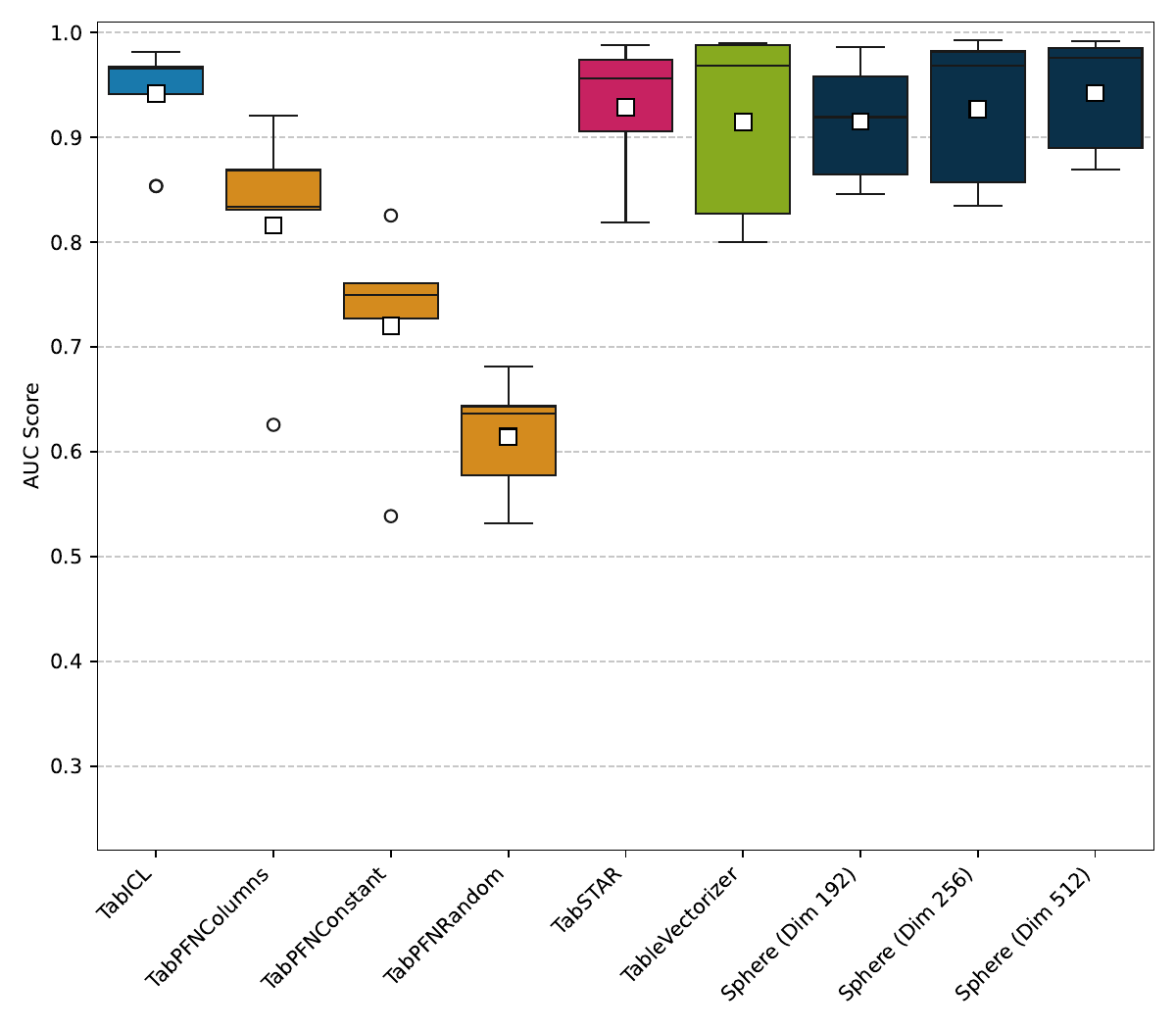}
    \caption{Distribution of the performance for multiclass classification evaluated via MLP-classifier.}
    \label{fig:boxplot_multiclass_mlp}
\end{figure}

\begin{figure}[p]
    \centering
    \includegraphics[width=0.8\linewidth]{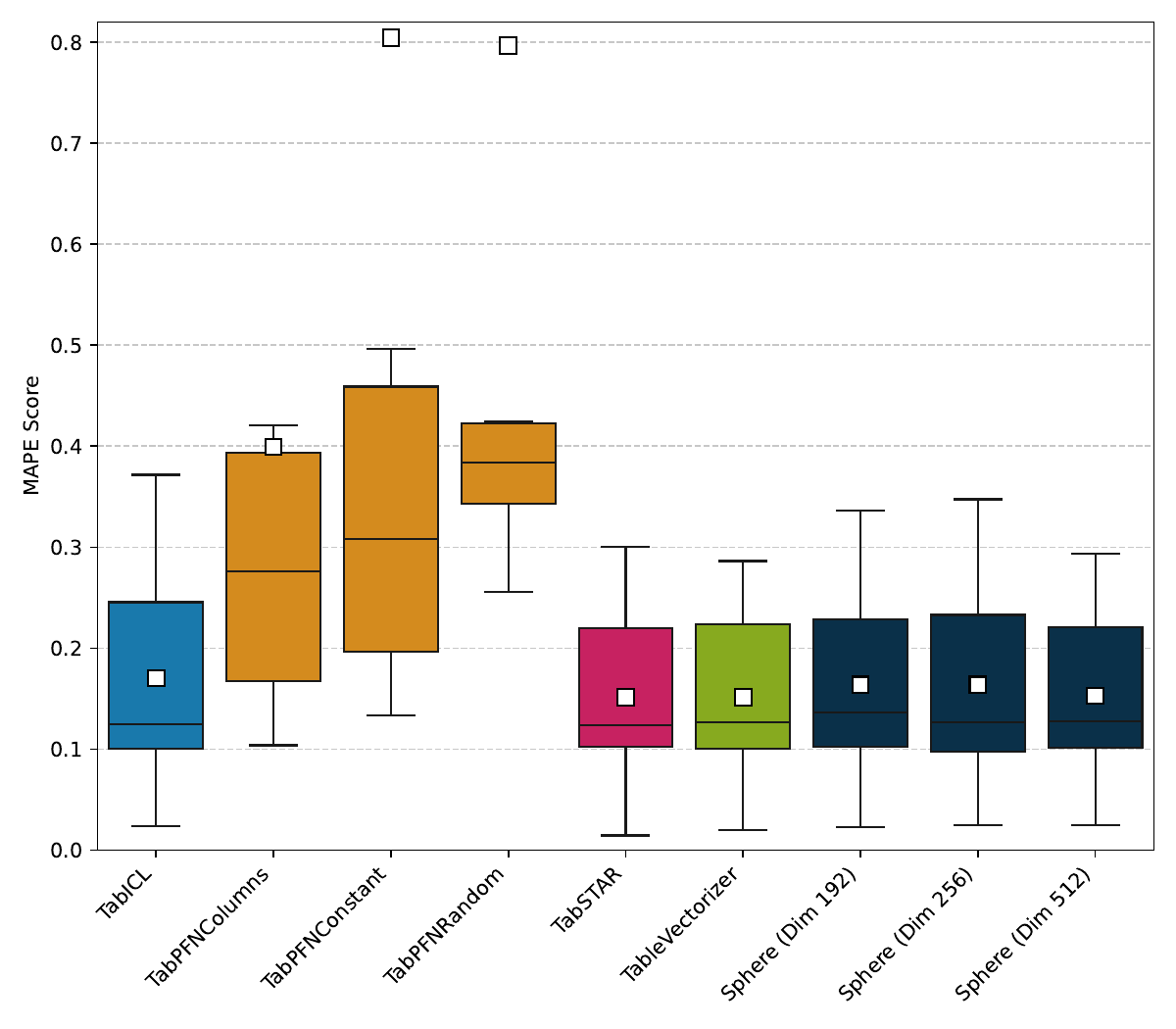}
    \caption{Distribution of the performance for regression evaluated via MLP-regressor. Note that for the TabPFN versions there are outliers that are not captured within the range of the vertical axis.}
    \label{fig:boxplot_regression_mlp}
\end{figure}

As we know the targets for the supervised tasks, for comparison we can apply the original tabular foundation models TabICLOriginal and TabPFNOriginal directly and compare the best scores: In binary classification with task-agnostic embeddings, the best averaged AUC score 0.82$\pm$ 0.09 is achieved by the sphere model with dimension 512 (via MLP-classifier). For comparison, TabICLOriginal achieves 0.85$\pm$0.10. For multiclass classification the best task-agnostic performance in terms of mean comes from TabICL and the 512-dimensional sphere model (both 0.94$\pm$0.05), whereas TabICLOriginal is slightly better with an AUC score 0.96$\pm$0.05. Finally 0.15$\pm$0.10 is the lowest (best) MAPE score obtained from several task-agnostic embedding models, TabPFNOriginal achieves 0.14$\pm$0.13. These results demonstrate that the overall performance for task-agnostic embedding models is comparable to the performance of original tabular foundation models, while the task-agnostic embeddings have the great advantage of being reusable for any other task.

Table~\ref{tab:time} refers to the average time needed to compute the embedding for one table row. Median, mean and standard deviation are calculated across all datasets. Note that the computation time is not directly comparable, since TabICL, TabPFN and TabSTAR use the GPU, whereas TableVectorizer and the sphere model run on the CPU. Nevertheless, TabICL, TableVectorizer and the sphere models are faster than TabPFN and TabSTAR. TabPFNColumns computes the embedding for each column and aggregates them by averaging, hence it is clear that TabPFNColumns requires significantly more computation time.

\begin{table}[h]
\centering
\begin{tabular}{@{$\;$}l@{$\;$}ccc@{$\;$}}
\hline
\textbf{Method} & \textbf{Median (ms)} & \textbf{Mean (ms)} & \textbf{Std (ms)}\\
\hline
TableVectorizer  &  \hspace*{0.185cm}0.17 &   \hspace*{0.37cm}0.24 &   \hspace*{0.37cm}0.28 \\
Sphere (Dim 192) &  \hspace*{0.185cm}0.08 &   \hspace*{0.37cm}0.16 &   \hspace*{0.37cm}0.21 \\
Sphere (Dim 256) &  \hspace*{0.185cm}0.09 &   \hspace*{0.37cm}0.18 &   \hspace*{0.37cm}0.25 \\
Sphere (Dim 512) &  \hspace*{0.185cm}0.17 &   \hspace*{0.37cm}0.32 &   \hspace*{0.37cm}0.45 \\
TabICL           &  \hspace*{0.185cm}0.16 &   \hspace*{0.37cm}0.35 &   \hspace*{0.37cm}0.50 \\
TabPFNColumns    &                28.06 &                153.61 &                528.91 \\
TabPFNConstant   &  \hspace*{0.185cm}1.49 &   \hspace*{0.37cm}2.96 &   \hspace*{0.37cm}7.44 \\
TabPFNRandom     &  \hspace*{0.185cm}1.97 &   \hspace*{0.37cm}4.15 &   \hspace*{0.37cm}7.73 \\
TabSTAR          &  \hspace*{0.185cm}1.06 &   \hspace*{0.37cm}1.93 &   \hspace*{0.37cm}3.34 \\
\hline
\end{tabular}
\caption{Average embedding time for one table row (one sample), compiled over all datasets.}
\label{tab:time}
\end{table}

\begin{figure*}[t]
    \centering
    \includegraphics[width=0.8\linewidth]{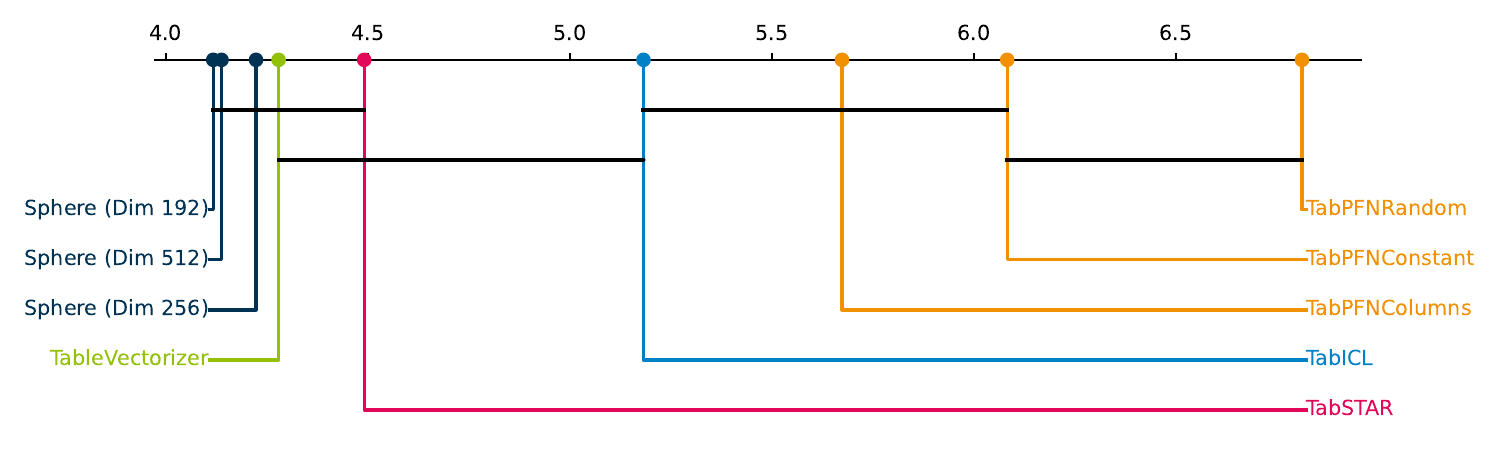}
    \caption{Critical difference diagram combined for all tasks.}
    \label{fig:critical_differences}
\end{figure*}

\section{Discussion}

Overall, Fig.~\ref{fig:critical_differences} presents a critical difference diagram combining every evaluation across all datasets. The diagram shows that TableVectorizer and the sphere models perform significantly better than TabICL and the TabPFN versions. Furthermore, TableVectorizer and TabSTAR outperform TabPFN. In summary, on the investigated benchmarks, classical feature engineering methods perform better than task-agnostic embeddings extracted from tabular foundation models. However, general ranking conclusions for the investigated embedding models that go beyond the evaluated datasets cannot be drawn from these results, as the number of datasets is limited. Furthermore, we extracted task-agnostic embeddings as described in Section~\ref{sec:embedding_models}. There might be ways to improve the extraction.

Regarding the embedding dimensionality, TableVectorizer constructs embeddings with the lowest dimension ($\leq$ 166), while TabPFN requires 192, TabSTAR 384 and TabICL even 512 as embedding dimension. A significant limitation of TableVectorizer is its variable embedding dimensionality, which is determined by the input dataset. This dataset-dependent dimensionality poses a challenge for applications requiring the joint embedding of diverse datasets into a unified representation space. In contrast, models like TabPFN and TabICL employ a fixed, but significantly larger embedding size across all datasets. The advantage of the simple sphere model is its arbitrary embedding dimension, while keeping a competitive or even superior performance. The choice of the dimension, e.g., 192, 256 or 512, does not seem to significantly influence the performance. 

Compared to original end-to-end tabular foundation models TabPFNOriginal and TabICLOriginal, the best task-agnostic embedding models (followed by KNN or MLP) do not exhibit significant performance loss. Moreover, the task-agnostic embeddings are reusable and could be applied in unsupervised settings such as similarity search, where the original end-to-end tabular foundation models cannot be applied a priori. On the other hand, these end-to-end models provide a direct prediction, while prediction based on task-agnostic embeddings may require intense hyperparameter training.

\section{Conclusion and Future Work}\label{sec:conclusion}
This study demonstrates that simple feature engineering methods like TableVectorizer or a simple sphere model achieve competitive performance for task-agnostic embeddings, while requiring substantially fewer computational resources. Prediction results based on task-agnostic embeddings are not significantly worse even in settings where sophisticated tabular foundation models can fully leverage their in-context learning capabilities. 

However, several open challenges remain for future work. First of all, our findings have to be confirmed on a wider range of datasets and more embedding models, e.g., TabPFNv2.5 \cite{grinsztajn2025tabpfn}, Mitra \cite{zhang2025mitra} and LimiX \cite{zhang2025limixunleashingstructureddatamodeling}. Second, although the embeddings are constructed in a task-independent way, their \emph{task-agnostic} quality should be further assessed considering additional downstream tasks and metrics.

Furthermore, current methods primarily capture statistical patterns. Recent works such as ConTextTab \cite{spinaci2025contexttab} and TabSTAR \cite{arazi2025tabstarf} demonstrate promising directions by incorporating textual (meta-)data. A comprehensive analysis for their task-agnostic embeddings incorporating text would be a further step toward multi-modal alignment, enabling tabular representations to be aligned with text or images for cross-modal retrieval and reasoning. The introduced sphere model has an advantage for aligning with other modalities, due to its arbitrary embedding dimension.

Large amounts of data occur subsequently and not as a closed dataset. The ingestion of new rows raises fundamental questions about context selection for in-context learners: Should entire datasets be re-embedded, or can a representative subset suffice? Distribution shift poses additional complications. The behavior of (in-context) embedding models needs to be investigated in such scenarios.

Extension to relational databases represents a natural progression, where tabular embedding models could serve as building blocks for relational models that learn representations across joined tables and foreign-key relationships. Shared representation spaces across diverse datasets are also worth investigation: Do models sufficiently differentiate between datasets when embedded jointly, or does cross-dataset contamination occur?

Addressing all these challenges will be essential for understanding (task-agnostic) tabular embeddings as a flexible, reusable tool in the broader machine learning ecosystem.

\clearpage
\bibliographystyle{plain}
\bibliography{literature}

\end{document}